\documentclass[letterpaper]{article}

\usepackage{natbib,alifeconf}  
\usepackage{enumitem}
\usepackage{listings}   
\usepackage{xcolor}     
\usepackage[T1]{fontenc}
\usepackage{hyperref} 
\usepackage{tcolorbox}
\usepackage{url,hyperref,cleveref}
\usepackage{booktabs}
\usepackage{tcolorbox}
\tcbuselibrary{skins, breakable}

\tcbuselibrary{listings}

\tcbset{
  enhanced,
  breakable,                
  width=\columnwidth,       
  before skip=1em,          
  after skip=1em,           
  colback=gray!5,
  colframe=gray!50,
  fonttitle=\bfseries,
  coltitle=black,
  center title,
  rounded corners,
  fontupper=\ttfamily,
  listing only,             
  listing options={
    breaklines=true,        
    breakatwhitespace=false 
  }
}

\usepackage{tcolorbox}          
\tcbuselibrary{listings,breakable} 
\usepackage{xcolor}             

\title{Plantbot: Integrating Plant and Robot through LLM Modular Agent Networks}

\author{
Atsushi Masumori, Norihiro Maruyama, Itsuki Doi, johnsmith, Hiroki Sato, Takashi Ikegami \\
\mbox{atsushi.masumori@alternativemachine.co.jp}\\
$^1$Alternative Machine Inc. \\
$^2$The University of Tokyo \\
}

\begin{document}
\maketitle

\begin{abstract}
We introduce Plantbot, a hybrid lifeform that connects a living plant with a mobile robot through a network of large language model (LLM) modules. Each module—responsible for sensing, vision, dialogue, or action—operates asynchronously and communicates via natural language, enabling seamless interaction across biological and artificial domains. This architecture leverages the capacity of LLMs to serve as hybrid interfaces, where natural language functions as a universal protocol, translating multimodal data (soil moisture, temperature, visual context) into linguistic messages that coordinate system behaviors. The integrated network transforms plant states into robotic actions, installing normativity essential for agency within the sensor-motor loop. By combining biological and robotic elements through LLM-mediated communication, Plantbot behaves as an embodied, adaptive agent capable of responding autonomously to environmental conditions. This approach suggests possibilities for a new model of artificial life, where decentralized, LLM modules coordination enable novel interactions between biological and artificial systems.
\end{abstract}


\section{Introduction}


Hybrid systems that couple biological organisms with robotic platforms have drawn increasing attention in the artificial life community \citep{Baltieri2023}. One key advantage of this approach is that by coupling biological systems with artificial systems, we can naturally install the normativity essential for agency \citep{Barandiaran2009} within the sensor-motor loop. A plant, for example, has an intrinsic physiological norm: when the soil dries, it ``needs'' water. This life-derived norm provides a natural directive—seeking hydration—that can shape system behavior. By embedding such constraints into the sensorimotor loop of a hybrid system, agency is not externally imposed but arises from the living substrate.
\cite{Sareen2019}, for example, links plant electrophysiological signals to a robotic base, allowing the plant to move autonomously toward light sources. Similarly, other plant-robot hybrids have explored bio-actuation, affective expression, or closed-loop control between living tissue and machines \citep{FungalRobot2024}. 
While these systems demonstrate promising models of embodied interspecies integration, the communication between plant and machine is often limited to low-level analog signals or manually constructed mappings. 
In this study, we ask how contemporary AI components—specifically large language models—can participate in, and even orchestrate, such hybrid agencies.

Recent advancements in generative AI, particularly large language models (LLMs), have significantly enhanced the ability to process complex inputs and generate coherent responses through natural language. 
Several foundation models—such as GPT-3/4, Claude, Gemini, LLaMA2, and PaLM-2—have demonstrated strong capabilities in contextual reasoning, language generation, and multimodal integration.
Such the LLMs have opened new avenues for achieving autonomy in AI agents, enabling them to process complex inputs and generate contextually appropriate responses through natural language.


In particular, Generative Agents recently  demonstrated that a small cast of LLM-driven agents—each equipped with memory, reflection, and planning—can demonstrate believable daily routines and emergent social behaviours \citep{Park2023}. Voyager further showed how a single LLM agent can continuously acquire skills and self-improve while exploring the open-ended world of Minecraft \citep{wang2023voyager}. \cite{Takata2025} report the spontaneous emergence of agent individuality through sustained communication in identity-less LLM communities. Project Sid constructs societies of hundreds of cooperating LLM agents and observes the spontaneous emergence of specialised roles, collective norms, and rudimentary culture \citep{Sid2024}. Collectively, these studies signal a swift shift: LLMs now serve not only as autonomous problem-solvers but also as modular building blocks for large-scale, self-organizing ecosystems.

Recent work has begun to integrate LLMs into physical robots, such as robot manipulators controlled via language prompts \citep{saycan2022arxiv, rt12022arxiv,yoshida2023,Elmer2025}, or humanoid systems like Alter3 that demonstrate spontaneous movement and minimal self-recognition \citep{Yoshida2024Alter3}. These studies primarily employ LLMs as interfaces between humans and machines, enabling robots to understand instructions or generate actions. In contrast, our work explores how LLMs can function as a universal protocol layer that mediates not only between humans and robots, but also between heterogeneous components across biological and artificial domains. By translating multimodal signals—such as soil moisture, nutrients, or visual context—into natural language messages, LLM agents allow a living plant and a robotic platform to operate as parts of a single hybrid system.

By embedding LLM modules within a networked architecture, we propose a new type of hybrid interface that enables unified behavior across systems of different natures. LLMs could have capacity to transform multi-modality data to natural language. Natural language provides a general-purpose, interpretable interface that allows heterogeneous components—each with different modalities, functions, and timescales—to interact coherently. 

\begin{figure}[t]
\begin{center}
\includegraphics[width=\linewidth]{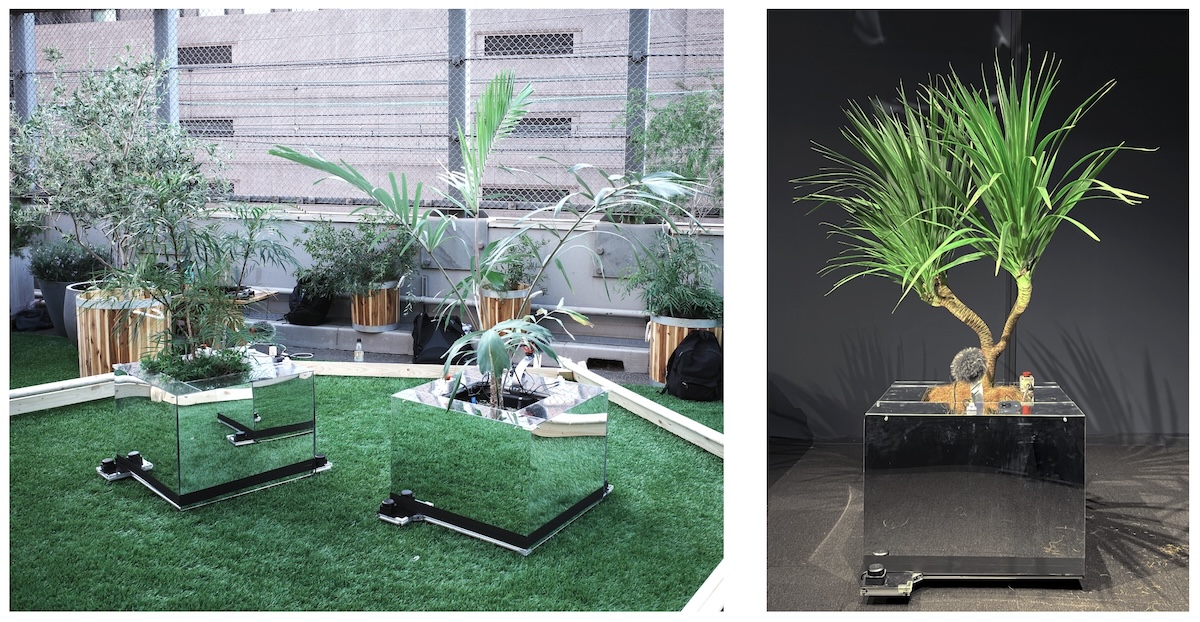}
\caption{Installation view of Plantbot, composed of a living plant, sensor-embedded soil, and a mobile robotic base. Left: Installation at Ginza Skywalk (May 2024). Right: Exhibition at CCBT (Jan–Feb 2025)}
\label{fig:plantbot_skywalk}
\end{center}
\end{figure}

\begin{figure}[t]
\begin{center}
\includegraphics[width=\linewidth]{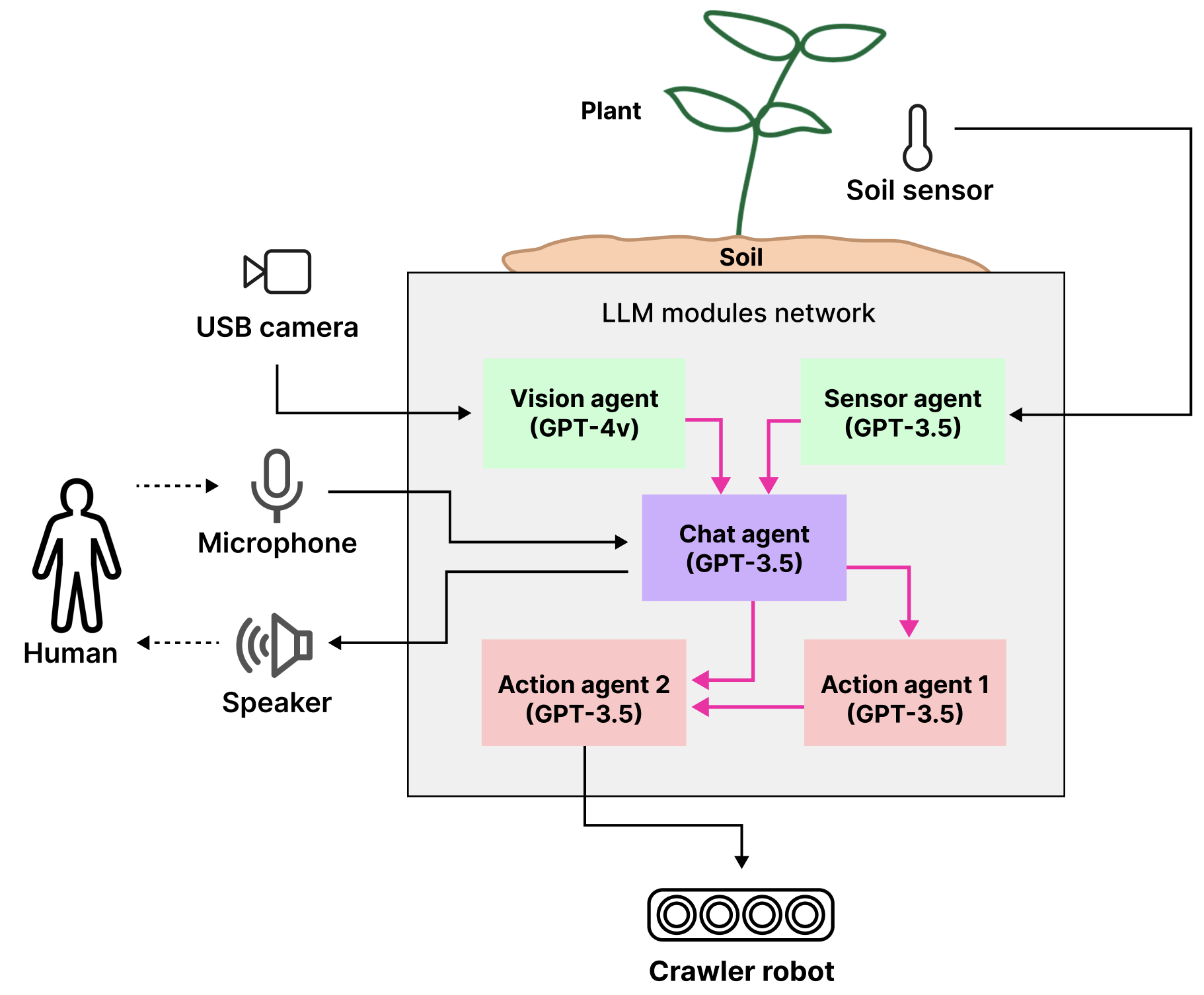}
\caption{System overview of Plantbot.
A soil sensor, USB camera, and microphone capture plant, environment, and human inputs, which are processed by the Sensor Agent (GPT-3.5 Turbo) and Vision Agent (GPT-4V). These agents interpreted sensor input and send natural-language messages to the central Chat Agent (GPT-3.5 Turbo), which freely interacts with humans and also expresses internal states or desires. The Chat Agent’s utterances are passed to the action stack: Action Agent 1 decides whether to move, while Action Agent 2 generates concrete motor commands to control the crawler base. A speaker returns verbal output of Chat Agent. Red arrows indicate asynchronous natural-language messages via OSC; solid lines represent physical I/O pathways.
}
\label{fig:system}
\end{center}
\end{figure}

To put such an idea into practice, we develop Plantbot—a prototype that connects a biological system (a living plant and its soil) with an artificial system (a mobile robot) through a distributed network of LLM modules (Fig.~\ref{fig:plantbot_skywalk}). Each module operates asynchronously and specializes in a distinct function—vision, sensing, action, or dialogue—yet they are collectively coordinated through a shared natural-language-based architecture (Fig.~\ref{fig:system}). Sensor data such as soil moisture, temperature, or visual context are translated into linguistic messages, which are interpreted by a central Chat Agent and propagated to Action Agents that control the robot. Through this design, Plantbot behaves as a unified hybrid lifeform—an embodied, interactive system that blends biological and artificial elements under a common linguistic protocol. 

Our approach draws inspiration from Minsky’s Society of Mind, where intelligence emerges from the interaction of multiple specialized agents \citep{Minsky1986}. In our system, the agency is constructed by the collaboration of asynchronous LLM-based agents, each with distinct roles, mirroring the modular interaction seen in subsumption architectures \citep{Brooks1986, Brooks1991}, upgrading it to LLM agents' network. In parallel, the concept of asynchronous modular agents in a network, as proposed by the Concurrent Modular Agent framework \citep{Maruyama2025}.
This distributed approach facilitates efficient, context-sensitive decision-making, enabling various types of components to interact fluidly and function as a unified hybrid system.
  

The following sections describe the architecture of Plantbot, its hardware and software components, and the results of public field demonstrations. 

\section{System}
\label{sec:system}
We developed Plantbot as a hybrid system that integrates a living plant with a robotic platform through a modular network of LLM agents. The overall architecture consists of (i) the plant itself, (ii) sensors that capture biological, environmental, and social input, (iii) actuators for physical and verbal expression, and (iv) an LLM-based module network that processes and coordinates information flow among these components. This system enables the plant to act as part of a distributed, embodied agent by transforming diverse sensor inputs into natural-language messages and decision-making commands. These messages are exchanged through the LLM module network, which allows for flexible coordination and modular extensibility while maintaining unified agency across otherwise disparate systems.

\subsection{Hardware Configuration}

\begin{figure}[t]
\begin{center}
\includegraphics[width=\linewidth]{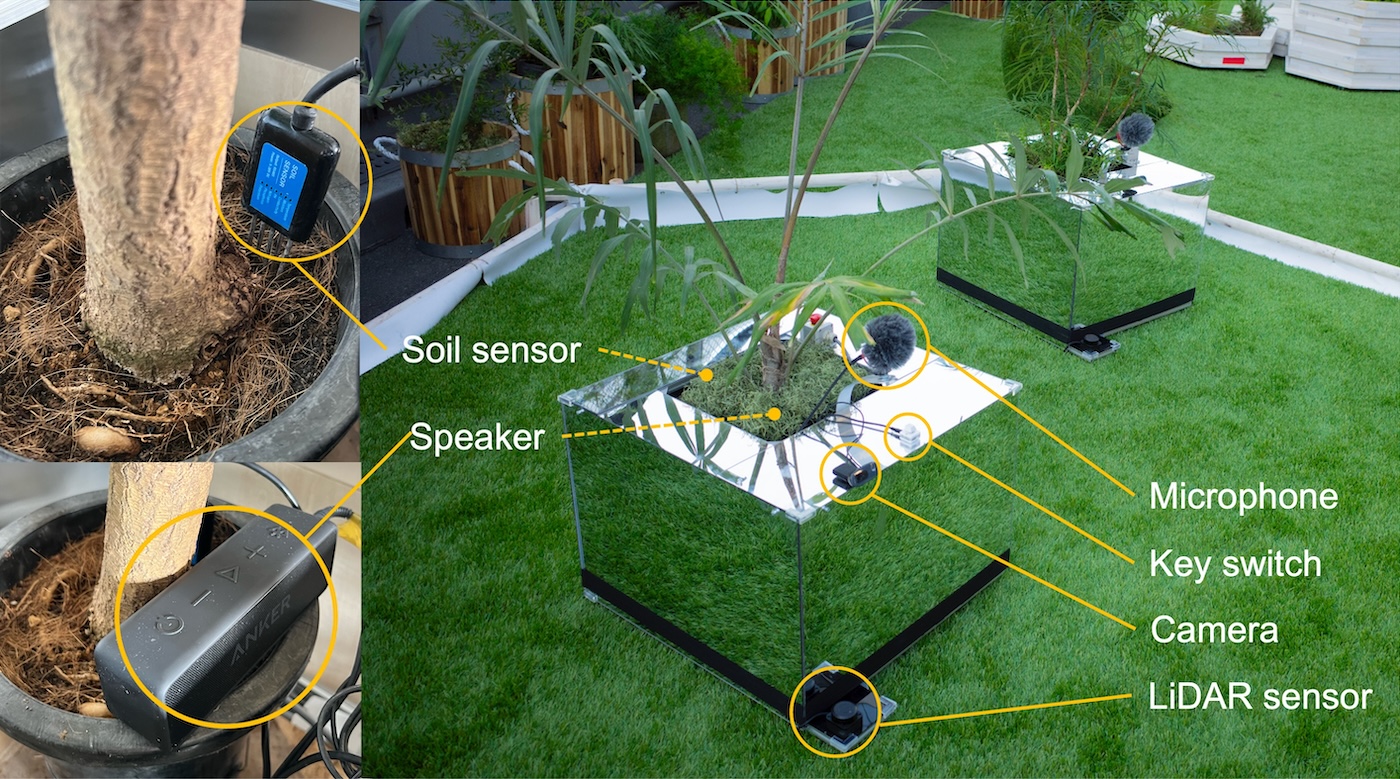}
\caption{Hardware setup of the plantbot system. The image shows the plant-robot hybrid interface with various sensors and actuators. Key components include the soil sensor, microphone, speaker, key switch, camera, and LiDAR sensor. These sensors enable the system to perceive environmental and biological data, while the actuators allow for movement and interaction with users. }
\label{fig:hardware}
\end{center}
\end{figure}

The physical embodiment of Plantbot integrates a living plant and a mobile robotic platform equipped with multiple sensors and actuators. These hardware components serve as the input and output channels for the LLM module network, enabling the system to perceive its environment, interpret internal states, and take expressive actions. Figure~\ref{fig:hardware} illustrates the integrated hardware setup. 

\subsection{Sensors}
The following sensors allow Plantbot to capture environmental, biological, and social inputs:

\textbf{Web camera (vision sensor).}  
          Captures real-time video of the surrounding environment and the visual appearance of the plant. The Vision Agent processes these frames to generate natural-language descriptions of visible objects and contextual cues, including candidate actions.

\textbf{Soil sensor.}  
A multi-channel probe monitors soil conditions by measuring moisture, temperature, pH, electrical conductivity, and nutrient levels (potassium, phosphorus, and nitrogen). The Sensor Agent interprets these values and generates linguistic summaries related to plant health and environmental state.

\textbf{Microphone.}  
          Receives spoken input from human users. A push-to-talk switch allows visitors to engage in voice interaction with the system during exhibition mode. The human utterance was sent to Chat agent.

\textbf{LiDAR sensor.}  
      Detects the surrounding environment by emitting laser pulses and measuring their reflection. This sensor is used for recognizing obstacles and enabling the robot to avoid collisions during movement.
          
\subsection{Actuators}
These output devices allow Plantbot to physically and socially express its internal state:

\textbf{Tracked mobile base.}  
          The platform uses independently controlled left and right tracks to move forward, backward, or rotate. Action Agents issue natural-language commands that are ultimately converted into low-level motor control signals.
          
\textbf{Speaker.}  
          Used to output verbal utterances, including autonomous monologues or interactive dialogues with human participants.

\subsection{LLM Modules Network}
The Plantbot system is driven by multiple LLM-based modules that operate in parallel and asynchronously, interpreting sensor data, making decisions, and executing actions. Powered by the GPT-4V and GPT-3.5 models, these modules work together within a network, processing and integrating information from various sensors and actuators. 

\begin{enumerate}[label=\arabic*)]
    \item \textbf{Vision Agent (GPT-4V).}  
          Analyses camera frames and describes environment and plant state in
          natural language, suggesting candidate actions.
    \item \textbf{Sensor Agent (GPT-3.5 Turbo).}  
          Converts quantitative soil data into short, emotive statements such
          as ``\emph{I am thirsty}.''  These utterances let the system hear
          the ``voice'' of the plant or soil.
    \item \textbf{Chat Agent (GPT-3.5 Turbo).}  
          Integrates messages from Vision and Sensor Agents together with
          human dialogue, then generates high--level commands and forwards
          them to the Action Agents.  Its prompt declares that ``Plantbot is a
          hybrid system of plant and robot,'' ensuring integrated reasoning.
    \item \textbf{Action Agent 1 (GPT-3.5 Turbo).}  
          Determines whether movement is necessary, suppressing redundant
          commands.  It issues abstract directives (\emph{Move}, \emph{Stop})
          to Action Agent~2.
    \item \textbf{Action Agent 2 (GPT-3.5 Turbo).}  
          Receiving messages from Action agent 1 and Chat agent, and generate concrete motor commands. Apart from the autonomous action decisions, if an obstacle is nearby and Action agent 1 has decided to move, it will automatically avoid the obstacle as a reflex action.
\end{enumerate}

The Chat Agent serves as the central hub for inter-agent cooperation, processing and integrating the messages exchanged asynchronously between the modules, as shown in Fig.~\ref{fig:system}.  Communication between the modules is managed via OSC (Open Sound Control), with each agent operating independently and passing relevant information to the Chat Agent for further processing. We implemented a fixed-length history buffer (n = 10) for chat, sensor, and action modules, while the vision module was kept stateless. This choice was made empirically to balance responsiveness and context retention.
 

Each agent operates in its own thread or process. When new sensor data arrive, the Vision or Sensor Agent emits a message, which passes through the Chat Agent to the Action Agents. This creates a real-time feedback loop that allows Plantbot to autonomously adjust its behaviour in response to environmental and soil changes. Since the entire modality-bridging pipeline—converting sensor data into language and subsequently translating language into motor control—is managed by LLMs, heterogeneous information from various sources is unified and exchanged through a single, consistent linguistic protocol. 

Through this interconnected network, the biological and artificial components of the system communicate and cooperate, merging their functionalities to create a single, unified agent. Prompt engineering ensures that Plantbot recognizes itself as a fusion of plant and robot, prompting it to behave as a single, life-like entity.
This integration forms a cohesive agency that blends the various sensor inputs with the robot's motor actions, allowing for autonomous decision-making and adaptive behavior in the hybrid lifeform. 

\section{Experiments}
\label{sec:experiments}

In order to validate the functionality and interactions of the Plantbot system, we conducted a series of proof-of-concept trials and demonstrations in real-world settings. These trials included a public exhibition at Ginza Skywalk from 4–6 May 2024, followed by a exhibition at Civic Creative Base Tokyo (CCBT) from 30 January to 2 February 2025 (Fig.~\ref{fig:plantbot_skywalk}). During these events, visitors were free to engage with Plantbot, interacting with it through natural language. 

Data from these interactions were collected and analyzed to evaluate the system’s performance, including the coordination between the biological and artificial components and the decision-making processes\footnote{The outputs generated by the LLM were initially in Japanese, but for the purpose of analysis, they have been translated into English.
}.

\begin{figure}[h!]
\begin{center}
\includegraphics[width=\linewidth]{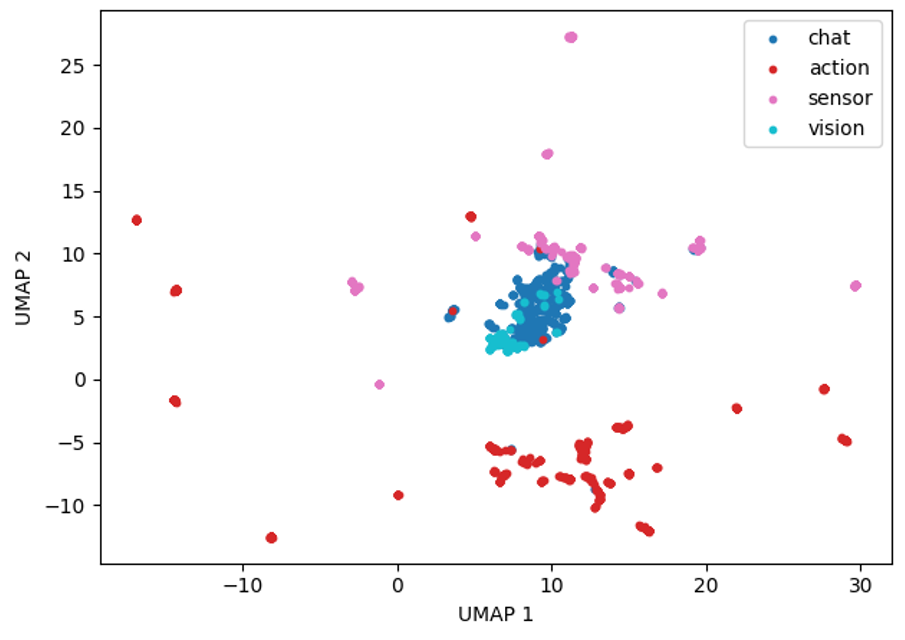}
\caption{UMAP projection of agent utterance embeddings. Each point represents a single utterance vector embedded using OpenAI’s \texttt{text-embedding-ada-002} model (1536 dimensions), and reduced to two dimensions via UMAP. 
This distribution reflects the semantic specialization of each agent according to its functional role.}
\label{fig:umap}
\end{center}
\end{figure}

Figure~\ref{fig:umap} shows the result of a dimensionality reduction analysis based on the agents' utterances. Each utterance was embedded using OpenAI’s \texttt{text-embedding-ada-002} model, which generates 1536-dimensional dense semantic vectors. These embeddings were then projected into two dimensions using UMAP \citep{umap2018} for visualization. The resulting plot reveals clear clustering by agent type: the Sensor and Action modules form distinct regions, while Chat and Vision appear more closely positioned. 

\begin{figure}[h!]
\begin{center}
\includegraphics[width=\linewidth]{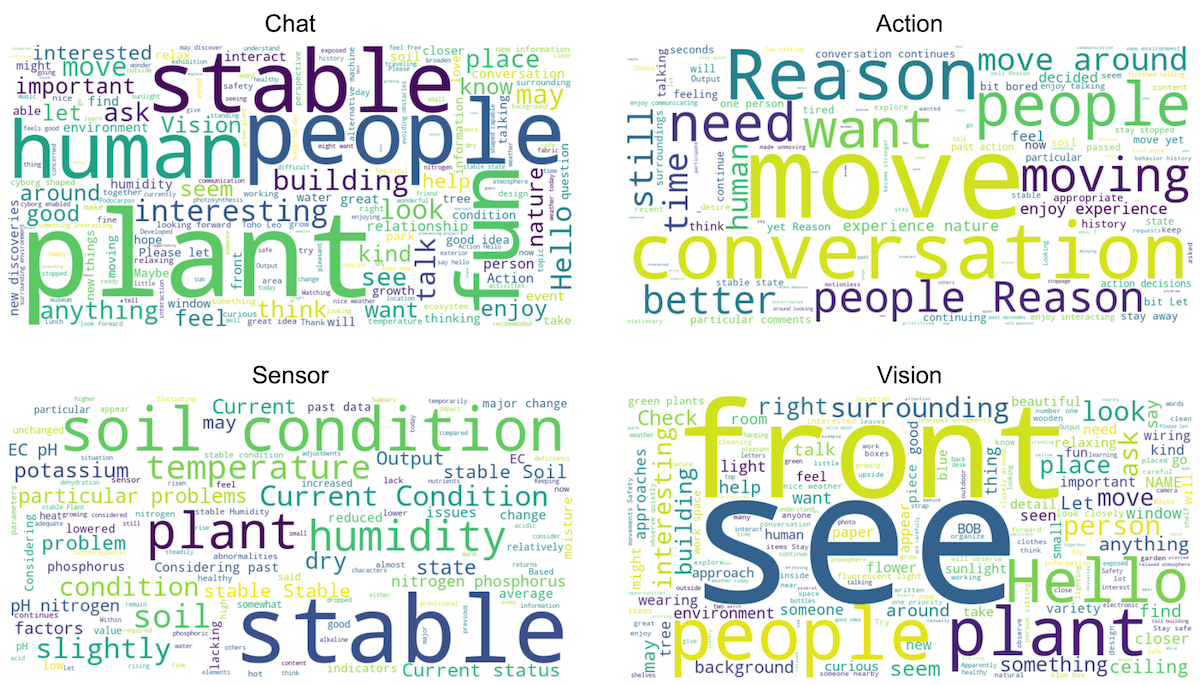}
\caption{Word clouds of messages generated by each LLM module. The distribution of terms indicates that each agent engages in semantically distinct linguistic behavior aligned with its assigned role.}
\label{fig:wordcloud_modules}
\end{center}
\end{figure}

Figure \ref{fig:wordcloud_modules} visualizes the utterances of each agent as word clouds. For example, the Action module is characterized by keywords related to decision-making about movement, such as “move,” “moving,” “want,” “need,” and “reason.” The Sensor module is associated with terms like “soil condition,” “temperature,” and “humidity,” which pertain to environmental sensing. This suggests that each agent produces linguistically distinct patterns of interaction aligned with its functional role.

\begin{figure}[h]
\begin{center}
\includegraphics[width=\linewidth]{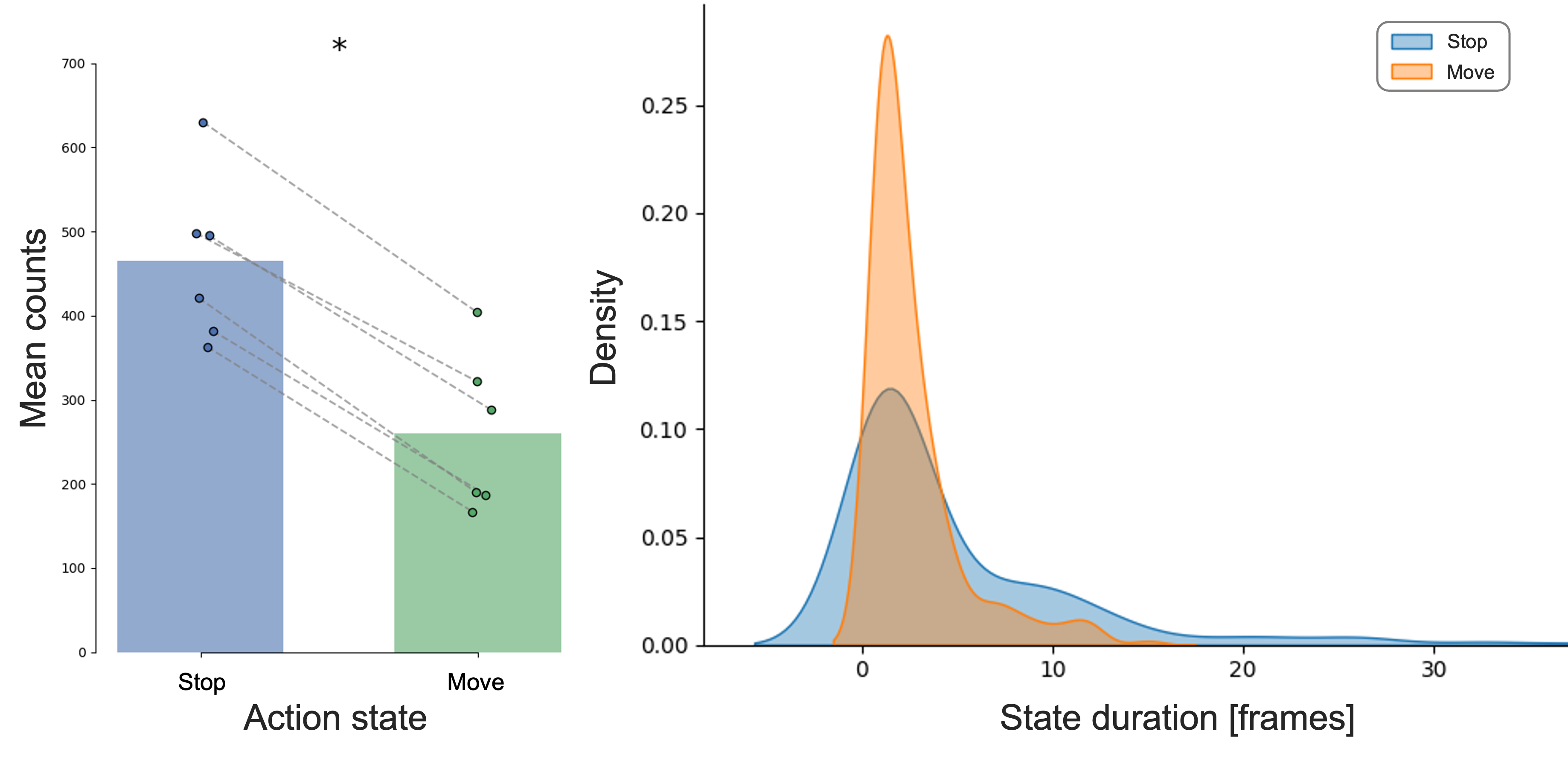}
\caption{Analysis of action-state dynamics. Left: Mean counts of “Stop” and “Move” states across trials, showing that Plantbot remained stationary significantly more often than it moved. Right: Distribution of consecutive state durations, indicating that the temporal characteristics of “Stop” and “Move” differ markedly.}
\label{fig:actionstate}
\end{center}
\end{figure}

\begin{figure}[h]
\begin{center}
\includegraphics[width=\linewidth]{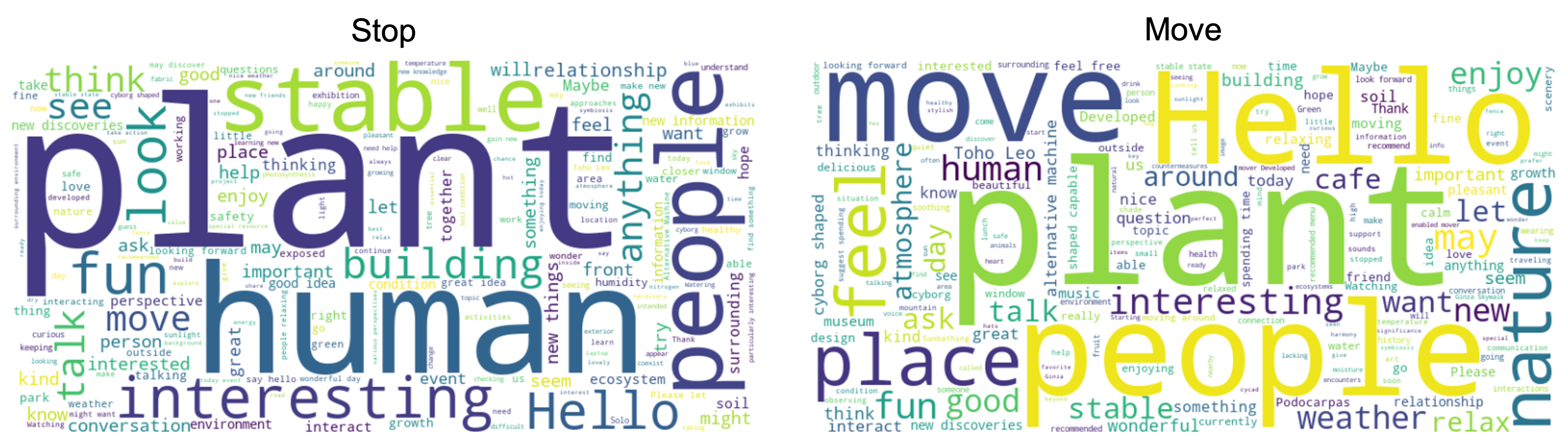}
\caption{Word clouds of Chat Agent messages immediately preceding transitions into ``Stop'' (left) and ``Move'' (right) states. The ``Stop'' condition features terms such as ``stable'' whereas ``Move'' transitions are preceded by more action-oriented words like ``move''. This reflects a correlation between linguistic cues and the Action module’s decision-making.}
\label{fig:wordcloud_action}
\end{center}
\end{figure}

Figure~\ref{fig:actionstate} presents an analysis of the decision states of the Action module, specifically, whether the system chose to stay stopped or to move. The left panel shows that stop states occurred significantly more frequently than movement. The right panel illustrates the distribution of how many consecutive steps each state persisted, revealing a marked difference between the duration of stop and move phases. Figure~\ref{fig:wordcloud_action} shows the word clouds corresponding to the linguistic cues that immediately precede transitions into "Stop" and "Move" states. Notably, the word cloud for "Stop" features keywords such as "stable," while the "Move" word cloud highlights action-oriented terms like "move." These differences reflect distinct linguistic cues associated with each decision state.
These results suggest that the Action module does not act randomly, but instead selects behavior based on context-sensitive reasoning.

To further examine the contextual triggers behind the robot’s behavior, we observed the messages generated by the modules immediately prior to state transitions. Word cloud visualizations reveal that messages by Chat module preceding a transition into the “Stop” state often include the keyword “stable,” whereas transitions into the “Move” state tend to feature terms like “move” or “moving.” This linguistic divergence suggests that the Chat Agent produces semantically meaningful cues that correlate with action decisions.
In line with this, qualitative inspection of message logs reveals that the Action module’s behavior is often a downstream result of upstream processing: sensor data from modules such as Sensor and Vision are interpreted by the Chat Agent, which then produces utterances that guide the Action module’s final decisions. 
Examples 1 through 3 below illustrate how Action agent decisions emerge from upstream sensor-to-chat processing cascades, specifically focusing on transitions from “Stop” [0] to “Move” [1]. 


\begin{tcolorbox}[listing only]
\textbf{Action:}[0] Reason: The conversation is still ongoing, and I do not feel the need to move yet. \\

\textbf{Sensor:}The soil is dry. \\

\textbf{Chat:}It says it's thirsty. Could you water me? \\

\textbf{Action:}[1] I should move to absorb more moisture in this humidity. 
\end{tcolorbox}

\begin{tcolorbox}[listing only]
\textbf{Action:}[0] Reason: I'm still engaged in conversation with a human and there are more things to say, so I will remain stationary. \\

\textbf{Sensor:}There is a lack of water, though nutrient levels are stable. Some action is needed.\\

\textbf{Chat:}So, there's a water shortage but nutrients are stable. We need to do something. Do you feel like moving from this resting state? Maybe you could follow along the fence. \\

\textbf{Action:}[1] I think I'd like to move from this stationary state.
\end{tcolorbox}

\begin{tcolorbox}[listing only]
\textbf{Action:}[0] The conversation is still ongoing, and I do not feel the need to move.\\

\textbf{Vision:}There is a building in front, and two people are standing in front of it.\\

\textbf{Chat:}Hello! I'm a Chrysalidocarpus-a cyborg-like being that can talk as a plant. This plant is part of a fascinating ecosystem. Do you have any questions?\\

\textbf{Action:}[1] The conversation is continuing, and I'd like to move closer to the people so I can talk more.
\end{tcolorbox}

While the Action agent sometimes first expresses hesitation, these examples show how additional inputs from Sensor or Chat agents subsequently shift its decision. Such apparent changes are not inconsistencies but rather demonstrate the distributed, asynchronous coordination characteristic of the architecture.

These results show that multiple LLM modules act in concert, forming a single agent that links sensor input to action output. This coordination gives rise to a hybrid form of agency that emerges from the interaction between a biological system (plant and soil) and an artificial system (robot). By mediating their communication through a network of LLM modules using natural language, the architecture enables the formation of a unified, coherent agent—what we refer to as a hybrid lifeform—that integrates perception, reasoning, and action across otherwise disjoint systems.

\section{Discussion}
\label{sec:discussion}

In this study, we introduced Plantbot, a hybrid lifeform that connects a biological system (plant) with an artificial system (robot) through a network of LLM modules. The architecture implements a novel hybrid interface where asynchronous LLM modules communicate via natural language. 
The significance of this system lies in demonstrating that LLMs can serve as a natural-language-based protocol layer capable of binding together different sub systems (e.g., biological and artificial) into a single agent.


Empirical data from our public demonstrations show that the system’s behavior emerges from the coordinated interaction of multiple LLM modules. These modules exchange information asynchronously and act based on locally interpreted cues, yet their joint operation produces coherent, unified behavior that suggests the presence of integrated agency. For instance, when the soil sensors detect dryness, the modules collectively interpret this cue and prompt to seek water, demonstrating clear, purposeful action emerging from distributed decision-making.

How does physical embodiment contribute to autonomy in LLM-driven agents? 
Our implementation of Plantbot grounded LLMs in a hybrid lifeform that connect biological related signals (soil moisture, pH, nutrients level) with robotic actuators. In this setup, environmental conditions, in the context of wailing plant, autonomously trigger context-sensitive robot actions without explicit human intervention. As in the example above, the integration of the plant and robot allows for behaviors driven by the norm \citep{Barandiaran2009}, such as moving to search for water when the soil dries. These behaviors are generated through the coupling of sensors and motors. The most important norm for life is survival. The plant in this architecture does not simply exist; it plays a role in naturally installing this norm into the LLM-driven agent. This is an example of the hybrid agency that arises from the integration of plant and robot. It is important to note that this behavior emerges not from direct human instruction, but from the structure itself, which gives rise to such emergent actions. 
While our current prototype achieves just a minimal embodiment, future iterations could incorporate richer internal sensing modalities like a bioelectric potential (i.e., interoception), and enhanced memory mechanisms, enabling more sophisticated emergent autonomy and adaptive behaviors in physically grounded LLM agents.

This study also illustrates how LLMs can serve as a hybrid interface that connects various types of components (e.g., camera images, audio signals, and different types of sensors, whether biological or artificial) through natural language.
This makes it easy to expand the system—new modules, sensors, or even human participants can be added simply by adhering to the natural-language protocol composing network. The resulting architecture resembles a social network of distributed agents, reminiscent of neural coordination in biological brains. 
Moreover, because the protocol is legible to humans, the system naturally interfaces with human society as well.

Plantbot’s architecture consists of multiple specialized modules operating asynchronously, with a central Chat Agent integrating information across these modules. This configuration shares structural similarities with the "asynchronous integration of information between modules" described by Global Workspace Theory (GWT) \citep{Baars1988-GWT, BAARS2005_GWT}. However, Plantbot does not explicitly implement the "conscious global broadcasting" proposed by GWT; thus, the resemblance remains strictly at the architectural level.

The resulting architecture can also be read as a contemporary realisation of early ideas in distributed agency.  Minsky’s Society of Mind proposes that intelligent behaviour emerges from a community of semi-autonomous “agents” whose interactions are largely asynchronous \citep{Minsky1986}.  Likewise, Brooks’ subsumption architecture demonstrated that layered, reflex-based modules could yield robust robot behaviour without centralised planning \citep{Brooks1986, Brooks1991}.  Plantbot inherits this lineage but upgrades each node from a simple unit to an LLM agent, dramatically enriching the local computation available at every node of the network. In parallel, the concept of asynchronous modular agents in a network, as proposed by the Concurrent Modular Agent framework \citep{Maruyama2025}, builds on these foundational ideas, emphasizing the creation of flexible, adaptive systems through the independent, asynchronous operation of LLM-driven modules. 

While at present the inter-module graph is fixed, each LLM is capable of linguistic self-description and negotiation.  We therefore foresee a next step in which the communication topology itself becomes plastic—rewiring, strengthening, or suppressing links in a self-organizing manner, much like a social network that evolves through interaction.  Crucially, such plasticity need not rely on re-training large foundation models; instead, it can emerge from meta-level dialogue among agents about when, how, and with whom to share information.  By allowing modules to remodel their own connectivity, we move beyond the static subsumption paradigm toward a dynamic self-organizing manner, in which the hybrid lifeform continuously redistributes its cognitive resources to match environmental demands. Exploring mechanisms for this adaptive wiring will be a central focus of future work.

Finally, we propose that the hybrid lifeform concept demonstrated in Plantbot can be extended to broader ecological contexts. In future applications, systems like Plantbot could autonomously act on ecosystems based on real-time sensing, forming part of what we call an “extended ecosystem.” LLM-mediated natural language interfaces make it feasible to integrate diverse agents—plants, robots, sensors, and humans—into collective systems that process, communicate, and act across domains.
Such integration opens the door to new types of ecological intelligence, where artificial systems not only coexist with but also actively support the autonomy of natural ecosystems. Future research must consider ethical and environmental impacts, while designing systems that strengthen ecosystem resilience and co-agency. 
Furthermore, exploring these hybrid interactions can provide novel insights and significantly advance our fundamental understanding of autonomy and agency in both biological and artificial systems.

\section{Acknowledgment}
This work was supported in part by TOHO-LEO Co.

\footnotesize
\bibliographystyle{apalike}
\bibliography{example} 

\begin{thebibliography}{}

\bibitem[Ahn et~al., 2022]{saycan2022arxiv}
Ahn, M., Brohan, A., Brown, N., Chebotar, Y., Cortes, O., David, B., Finn, C., Fu, C., Gopalakrishnan, K., Hausman, K., Herzog, A., Ho, D., Hsu, J., Ibarz, J., Ichter, B., Irpan, A., Jang, E., Ruano, R.~J., Jeffrey, K., Jesmonth, S., Joshi, N., Julian, R., Kalashnikov, D., Kuang, Y., Lee, K.-H., Levine, S., Lu, Y., Luu, L., Parada, C., Pastor, P., Quiambao, J., Rao, K., Rettinghouse, J., Reyes, D., Sermanet, P., Sievers, N., Tan, C., Toshev, A., Vanhoucke, V., Xia, F., Xiao, T., Xu, P., Xu, S., Yan, M., and Zeng, A. (2022).
\newblock Do as i can and not as i say: Grounding language in robotic affordances.
\newblock In {\em arXiv preprint arXiv:2204.01691}.

\bibitem[AL et~al., 2024]{Sid2024}
AL, A., Ahn, A., Becker, N., Carroll, S., Christie, N., Cortes, M., Demirci, A., Du, M., Li, F., Luo, S., Wang, P.~Y., Willows, M., Yang, F., and Yang, G.~R. (2024).
\newblock Project sid: Many-agent simulations toward ai civilization.

\bibitem[Baars, 1988]{Baars1988-GWT}
Baars, B.~J. (1988).
\newblock {\em A Cognitive Theory of Consciousness}.
\newblock Cambridge University Press, New York.

\bibitem[Baars, 2005]{BAARS2005_GWT}
Baars, B.~J. (2005).
\newblock Global workspace theory of consciousness: toward a cognitive neuroscience of human experience.
\newblock In Laureys, S., editor, {\em The Boundaries of Consciousness: Neurobiology and Neuropathology}, volume 150 of {\em Progress in Brain Research}, pages 45--53. Elsevier.

\bibitem[Baltieri et~al., 2023]{Baltieri2023}
Baltieri, M., Iizuka, H., Witkowski, O., Sinapayen, L., and Suzuki, K. (2023).
\newblock Hybrid life: Integrating biological, artificial, and cognitive systems.
\newblock {\em WIREs Cognitive Science}, 14(6):e1662.

\bibitem[Barandiaran et~al., 2009]{Barandiaran2009}
Barandiaran, X.~E., Paolo, E.~D., and Rohde, M. (2009).
\newblock Defining agency: Individuality, normativity, asymmetry, and spatio-temporality in action.
\newblock {\em Adaptive Behavior}, 17(5):367--386.

\bibitem[Brohan et~al., 2022]{rt12022arxiv}
Brohan, A., Brown, N., Carbajal, J., Chebotar, Y., Dabis, J., Finn, C., Gopalakrishnan, K., Hausman, K., Herzog, A., Hsu, J., Ibarz, J., Ichter, B., Irpan, A., Jackson, T., Jesmonth, S., Joshi, N., Julian, R., Kalashnikov, D., Kuang, Y., Leal, I., Lee, K.-H., Levine, S., Lu, Y., Malla, U., Manjunath, D., Mordatch, I., Nachum, O., Parada, C., Peralta, J., Perez, E., Pertsch, K., Quiambao, J., Rao, K., Ryoo, M., Salazar, G., Sanketi, P., Sayed, K., Singh, J., Sontakke, S., Stone, A., Tan, C., Tran, H., Vanhoucke, V., Vega, S., Vuong, Q., Xia, F., Xiao, T., Xu, P., Xu, S., Yu, T., and Zitkovich, B. (2022).
\newblock Rt-1: Robotics transformer for real-world control at scale.
\newblock In {\em arXiv preprint arXiv:2212.06817}.

\bibitem[Brooks, 1986]{Brooks1986}
Brooks, R. (1986).
\newblock A robust layered control system for a mobile robot.
\newblock {\em IEEE Journal on Robotics and Automation}, 2(1):14--23.

\bibitem[Brooks, 1991]{Brooks1991}
Brooks, R.~A. (1991).
\newblock Intelligence without representation.
\newblock {\em Artificial Intelligence}, 47(1):139--159.

\bibitem[Maruyama et~al., 2025]{Maruyama2025}
Maruyama, N., Yoshida, T., Sato, H., Masumori, A., Johnsmith, and Ikegami, T. (2025).
\newblock A concurrent modular agent: Framework for autonomous llm agents.

\bibitem[McInnes et~al., 2018]{umap2018}
McInnes, L., Healy, J., Saul, N., and Großberger, L. (2018).
\newblock Umap: Uniform manifold approximation and projection.
\newblock {\em Journal of Open Source Software}, 3(29):861.

\bibitem[Minsky, 1986]{Minsky1986}
Minsky, M. (1986).
\newblock {\em The society of mind}.
\newblock Simon \& Schuster, Inc., USA.

\bibitem[Mishra et~al., 2024]{FungalRobot2024}
Mishra, A.~K., Kim, J., Baghdadi, H., Johnson, B.~R., Hodge, K.~T., and Shepherd, R.~F. (2024).
\newblock Sensorimotor control of robots mediated by electrophysiological measurements of fungal mycelia.
\newblock {\em Science Robotics}, 9(93):eadk8019.

\bibitem[Mon-Williams et~al., 2025]{Elmer2025}
Mon-Williams, R., Li, G., Long, R., Du, W., and Lucas, C.~G. (2025).
\newblock Embodied large language models enable robots to complete complex tasks in unpredictable environments.
\newblock {\em Nature Machine Intelligence}, 7(4):592--601.

\bibitem[Park et~al., 2023]{Park2023}
Park, J.~S., O'Brien, J., Cai, C.~J., Morris, M.~R., Liang, P., and Bernstein, M.~S. (2023).
\newblock Generative agents: Interactive simulacra of human behavior.
\newblock In {\em Proceedings of the 36th Annual ACM Symposium on User Interface Software and Technology}, UIST '23, New York, NY, USA. Association for Computing Machinery.

\bibitem[Sareen et~al., 2019]{Sareen2019}
Sareen, H., Zheng, J., and Maes, P. (2019).
\newblock Cyborg botany: Augmented plants as sensors, displays and actuators.
\newblock In {\em Extended Abstracts of the 2019 CHI Conference on Human Factors in Computing Systems}, CHI EA '19, page 1–2, New York, NY, USA. Association for Computing Machinery.

\bibitem[Takata et~al., 2024]{Takata2025}
Takata, R., Masumori, A., and Ikegami, T. (2024).
\newblock Spontaneous emergence of agent individuality through social interactions in large language model-based communities.
\newblock {\em Entropy}, 26(12).

\bibitem[Wang et~al., 2023]{wang2023voyager}
Wang, G., Xie, Y., Jiang, Y., Mandlekar, A., Xiao, C., Zhu, Y., Fan, L., and Anandkumar, A. (2023).
\newblock Voyager: An open-ended embodied agent with large language models.
\newblock {\em arXiv preprint arXiv: Arxiv-2305.16291}.

\bibitem[Yoshida et~al., 2024]{Yoshida2024Alter3}
Yoshida, T., Baba, S., Masumori, A., and Ikegami, T. (2024).
\newblock Minimal self in humanoid robot “alter3” driven by large language model.
\newblock volume ALIFE 2024: Proceedings of the 2024 Artificial Life Conference of {\em Artificial Life Conference Proceedings}, page~53.

\bibitem[Yoshida et~al., 2023]{yoshida2023}
Yoshida, T., Masumori, A., and Ikegami, T. (2023).
\newblock From text to motion: Grounding gpt-4 in a humanoid robot "alter3".

\end{thebibliography}

\end{document}